\documentclass[11pt]{article}

\usepackage[final]{acl}

\usepackage{times}
\usepackage{latexsym}

\usepackage[T1]{fontenc}

\usepackage[utf8]{inputenc}

\usepackage{microtype}

\usepackage{inconsolata}

\usepackage{graphicx}

\usepackage{amsmath}
\usepackage{amssymb}
\usepackage{mathtools}
\usepackage{algorithm}
\usepackage{algpseudocode}
\usepackage{makecell}

\usepackage{booktabs}
\usepackage{multirow}

\title{Spexis: Speculative and Lookahead Scheduling for LLM Inference}

\author{
 \textbf{Hyungyu Jung\textsuperscript{1}},
 \textbf{Jaehyeok Yu\textsuperscript{1}},
 \textbf{Hoonseo Choi\textsuperscript{2}}, 
 \textbf{Sungkyun Kim\textsuperscript{2}}, 
 \textbf{Jinho Lee\textsuperscript{1}},
 \textbf{Jiwon Seo\textsuperscript{1}}
\\
\\
 \textsuperscript{1}Seoul National University,
 \textsuperscript{2}Hanyang University
}

\begin{document}
\maketitle
\begingroup
\renewcommand{\thefootnote}{}
\footnotetext{
  \textbf{Correspondence:} Jiwon Seo <\href{mailto:seojiwon@snu.ac.kr}{\texttt{seojiwon@snu.ac.kr}}>
}
\endgroup

\begin{abstract}
Spexis is a multi-GPU LLM inference framework that improves the efficiency of pipeline and tensor parallelism through speculative parallelism. Rather than using speculative decoding only to accelerate token generation, Spexis runs speculation in parallel with normal execution, introducing a new parallelism axis without increasing KV-cache memory usage. This improves memory efficiency and helps mitigate the bottlenecks of multi-GPU inference.

Spexis further uses lookahead scheduling to predict speculation quality and future memory pressure, allowing it to reduce wasted speculation, KV-cache eviction, and recomputation. Built on top of vLLM, Spexis largely improves serving performance across a range of GPU configurations, achieving speedups of up to 34\% over a baseline that uses the optimal combination of pipeline and tensor parallelism. Spexis's source code is publicly available at https://github.com/mlsys-seo/spexis.
\end{abstract}

\section{Introduction}
LLMs are compute-intensive and are thus typically served on GPUs or other hardware accelerators. Recent models have also grown significantly in parameter size and KV cache memory demand, requiring multiple GPUs for their execution.

For multi-GPU inference, model parallelism is widely used, notably tensor parallelism (TP) and pipeline parallelism (PP). However, both face performance bottlenecks: TP incurs synchronization overhead, while PP suffers from reduced batch sizes. The problem becomes worse as more GPUs are used. In PP, adding more pipeline stages increases memory pressure from micro-batches, while in TP, using more GPUs incurs higher communication overhead.

We propose Spexis, a scheduling technique that extends speculative decoding for scalable LLM inference. Spexis introduces speculative parallelism as a new scheduling primitive and executes speculation in parallel with normal execution. This design provides additional parallelism without increasing KV cache memory usage, making multi-GPU execution more memory-efficient.

To further improve memory efficiency, Spexis uses prediction-based scheduling to estimate future memory pressure and determine when to run new requests. This helps reduce eviction and recomputation overhead while improving batch efficiency.

Our evaluation shows that Spexis achieves up to a 34\% speedup. Our contributions are as follows.

\noindent {\bf Speculative-parallel scheduling.}
We introduce speculative-parallel (SP) scheduling, a new parallel execution axis that complements pipeline and tensor parallelism. We further improve it with selective speculation and remaining-length prediction.

\noindent{\bf Analytic cost model.}
We develop cost models for PP, TP, and SP to efficiently combine these methods. We also develop an eviction-aware model that captures recomputation costs to determine when to schedule incoming requests for higher throughput.

\noindent{\bf Implementation and evaluation.}
We implement Spexis on top of vLLM, a widely used LLM inference system. Across NVIDIA A40, RTX PRO 6000, H100, A100 and L40S GPUs, Spexis achieves speedups of up to 34\%.

The rest of the paper is organized as follows. Section~\ref{sec:relatedwork} reviews related work. Section~\ref{sec:motivation} presents the motivation for our work. Section~\ref{sec:spexis} introduces Spexis and speculative parallelism. Section~\ref{sec:lookahead} describes our lookahead scheduling optimization. Section~\ref{sec:implementation} presents our prototype implementation. Section~\ref{sec:eval} evaluates Spexis and Section~\ref{sec:conclusion} concludes.

\section{Related Work}
\label{sec:relatedwork}
\paragraph{Scheduling for single-node LLM inference.}
The unique execution characteristics of LLMs (particularly the distinct prefill and decode phases and highly variable sequence lengths) make it challenging to fully utilize computational resources. Prior work has addressed these challenges, including low decode utilization~\cite{orca}, fluctuating prefill overhead~\cite{sarathi}, and KV cache memory inefficiency~\cite{vllm}. While these approaches significantly improve performance in single-node deployments, they do not address the additional bottlenecks on distributing LLM inference across multiple GPUs or nodes.

\paragraph{Parallelism strategy for distributed inference.} 
When an LLM exceeds the memory capacity of a single GPU, it must be distributed across multiple GPUs. Tensor parallelism (TP)~\cite{megatron-lm} partitions the weight matrices of each Transformer layer across GPUs for concurrent execution. In multi-node deployments, pipeline parallelism (PP)~\cite{gpipe} is commonly used to partition model layers into multiple stages. Because TP and PP suffer from scalability limitations due to synchronization overhead and memory bottlenecks, respectively, hybrid PP+TP configurations have become standard in practice~\cite{vllm-doc}. However, it still does not fully eliminate the bottlenecks of PP and TP. Spexis mitigates these bottlenecks with its memory-efficient speculative-parallel execution.

\paragraph{Scheduling with length prediction.}  
In LLMs, KV cache memory grows proportionally with sequence length, yet total available memory is fixed and output length is unknown, making memory-aware scheduling inherently challenging. To mitigate this, prior work has explored generation length prediction to improve scheduling efficiency. S3~\cite{s3} enables larger batch sizes, LTR and TRAIL~\cite{ltr, trail} reduce average latency by prioritizing shorter requests in the scheduling queue. In contrast, our work uses predicted length information to reduce KV-cache eviction overhead and improve serving throughput. When the predicted lengths indicate that admitting a new request would exceed memory capacity, its prefill can be deferred. To guide this decision, we develop a cost model that explicitly balances the delay of new requests against the recomputation cost of evicted ones.

\paragraph{Speculative decoding.} Speculative decoding has been proposed and extensively studied to overcome the autoregressive nature of transformer-based models~\cite{sd}. In speculative decoding, a lightweight draft model predicts multiple tokens ahead, and the base model verifies them in parallel to accelerate generation. Various self-speculative approaches have been explored to eliminate the overhead of maintaining a separate draft model~\cite{draft&verify, layerskip, kangaroo}. For example, Kangaroo~\cite{kangaroo} reuses the early layers of the base model as a draft model while the remaining layers verify the drafted tokens. We adopt a similar self-speculative approach and apply speculative decoding to pipeline-parallel execution, effectively filling the pipeline idle time that causes the bubble problem in PP.

\section{Motivation: Bottleneck in LLM Serving}
\label{sec:motivation}

LLM serving often requires multiple GPUs due to large parameters and KV cache memory demand. Multi-GPU inference commonly uses pipeline parallelism (PP) or tensor parallelism (TP), but both face scalability bottlenecks: PP from memory pressure and TP from communication overhead.

\begin{figure}[t]
    \centering
    \includegraphics[width=1\linewidth]{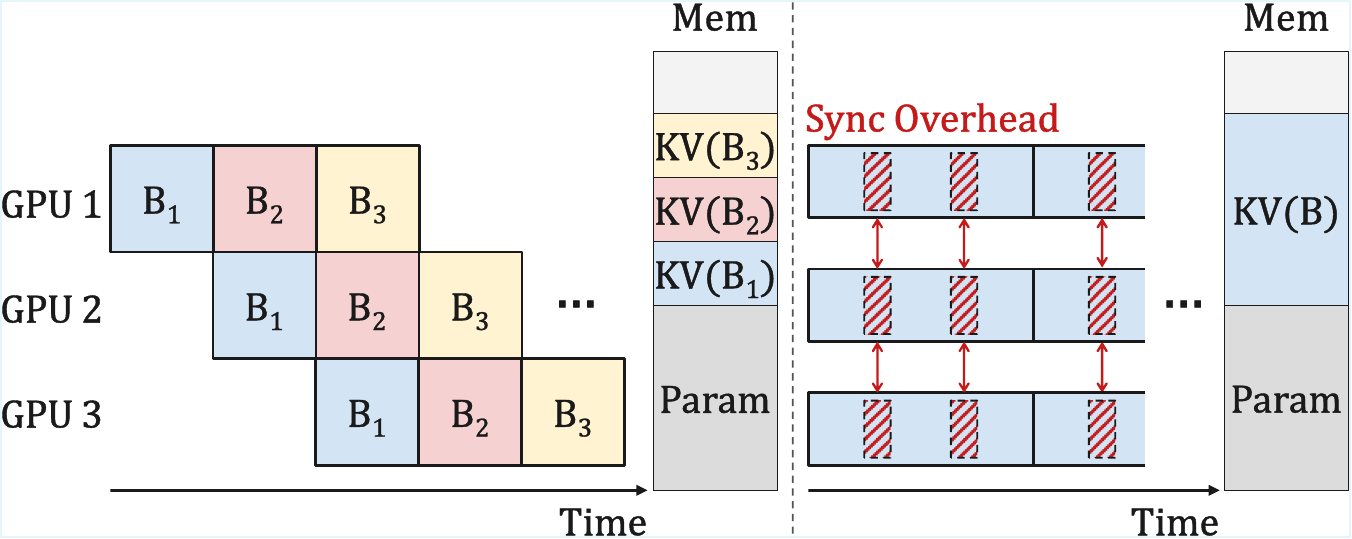}
    \caption{Execution timeline and memory usage of pipeline parallel execution (left) and tensor parallel execution (right) using three GPUs.}
    \label{fig:pp-tp-bottleneck}
\end{figure}

\noindent
{\bf Memory capacity bottleneck.} In pipeline-parallel execution, consecutive layers are grouped into stages and assigned across GPUs. To reduce pipeline bubbles caused by inter-stage dependencies, PP uses micro-batching, as illustrated in Figure~\ref{fig:pp-tp-bottleneck}. The main bottleneck in PP is the KV cache memory used by the micro-batches. Because different micro-batches process different sequences, each micro-batch requires its own KV cache entries, as shown in the figure. As a result, the size of each micro-batch must be much smaller than the batch size achievable without micro-batching. These small micro-batch sizes lead to low GPU utilization and, consequently, lower throughput.

\noindent {\bf Communication bottleneck.} In tensor-parallel execution, each layer and its weight parameters are partitioned across multiple GPUs. After each partitioned computation, the partial outputs are synchronized across the GPUs. For LLMs, this synchronization is required twice per transformer layer. As shown in Figure~\ref{fig:pp-tp-bottleneck}, GPUs remain idle during these synchronization steps, degrading overall resource utilization. High-bandwidth interconnects such as NVLink can mitigate this overhead, but the bottleneck is not fully eliminated. In particular, within a node, communication bandwidth is often non-uniform across GPU pairs, and all-reduce collectives across the local NVLink-connected GPU domain can still become a bottleneck.

These bottlenecks become worse as the degree of PP and TP increases. As a result, simply increasing model parallelism does not necessarily improve serving throughput and may even reduce it. This motivates a new parallel execution strategy that alleviates both bottlenecks while retaining the scalability benefits of multi-GPU execution.

\begin{figure}
    \centering
    \includegraphics[width=1\linewidth]{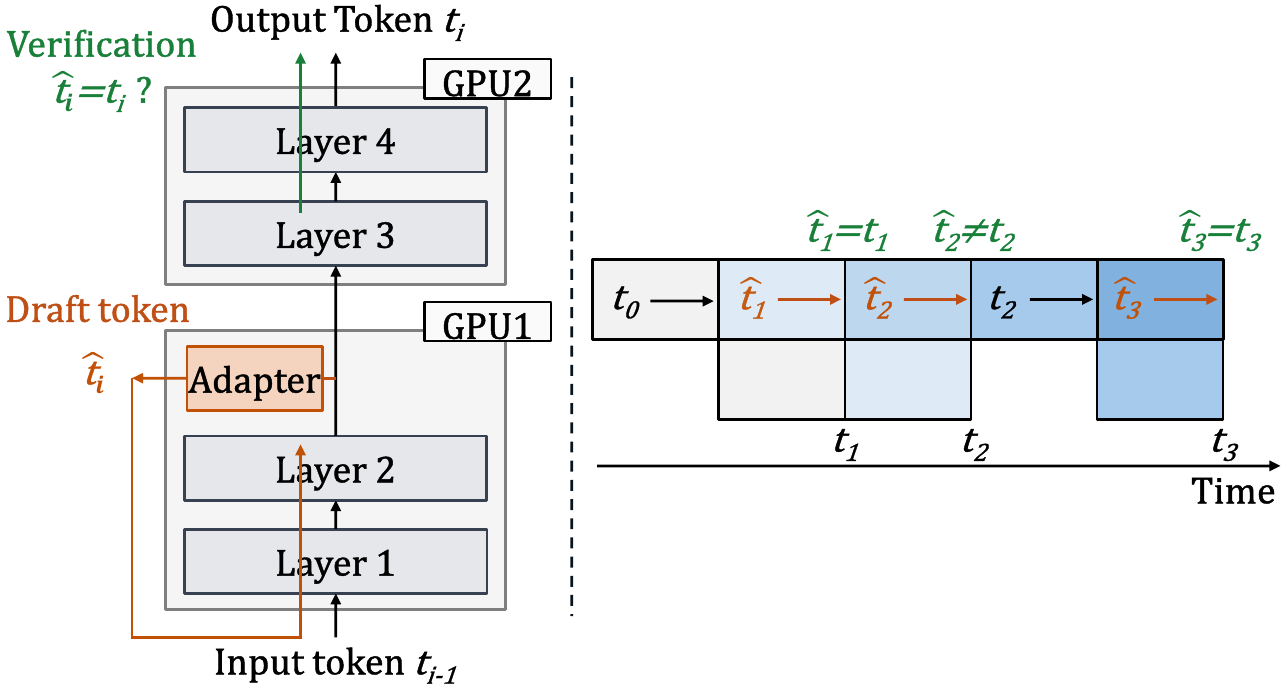}
    \caption{Overview of the self-speculative model and its execution timeline in Spexis under a two-stage pipeline. The timeline shows the execution of a single request: orange arrows indicate speculative execution, and black arrows indicate normal execution after speculation rejection.}
    \label{fig:spexis-overview}
\end{figure}

\section{Spexis: Speculative-Parallel Execution}
\label{sec:spexis}

To mitigate the bottlenecks of pipeline and tensor parallelism, we introduce a new scheduling axis, speculative parallelism (SP). By jointly optimizing pipeline parallelism (PP), tensor parallelism (TP), and speculative parallelism (SP), Spexis effectively minimizes these bottlenecks. 

\noindent 
{\bf Self-speculation and parallel scheduling.}
Self-speculative models use their earlier layers for drafting and the rest layers for verifying drafted tokens as part of their normal execution path. Similar to conventional speculative decoding, self-speculation drafts multiple tokens and then verify them jointly.

In Spexis, we execute speculation in parallel with normal model execution (i.e., verification) while running the entire model in a pipeline- and/or tensor-parallel manner. As a simple example, Figure~\ref{fig:spexis-overview} shows a self-speculative model and its execution in Spexis using a two-stage pipeline on two GPUs. After generating draft tokens from the first stage, Spexis immediately begins speculative execution for those tokens in the first stage, in parallel with their verification in the second stage. 

Because speculative execution does not require a separate KV cache and instead shares the cache with normal execution, it supports larger batch sizes, leading to higher throughput. In comparison, PP's micro-batching uses separate KV caches for each micro-batch, which incurs the memory-capacity bottleneck discussed in Section 3. Furthermore, when SP is applied, the required degree of TP can be reduced, which in turn lowers its communication overhead.

The effectiveness of SP scheduling largely depends on the acceptance rate of draft tokens. In speculative decoding, the first few draft tokens, especially in self-speculative models, tend to have relatively high acceptance rates. For example, prior work reports a maximum acceptance rate of 85\% for the first draft token in EAGLE~\cite{eagle}, while our evaluation shows a maximum acceptance rate of 78\% for the first draft token. These high early-token acceptance rates make SP practical and allow it to be combined effectively with TP and PP to optimize LLM inference.

\begin{figure}
    \centering
    \includegraphics[width=1\linewidth]{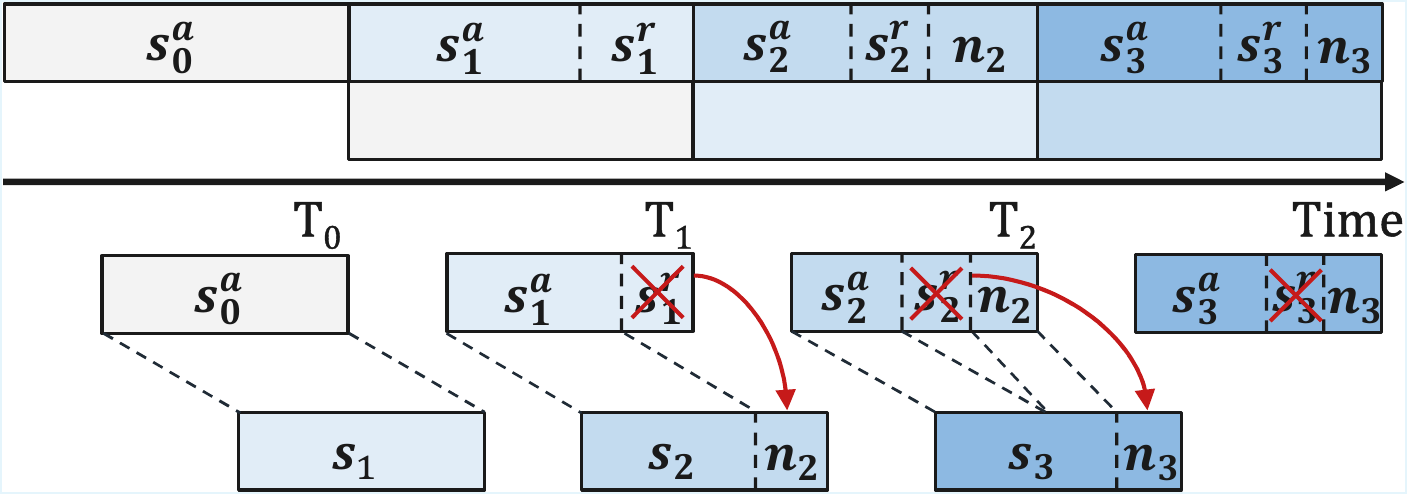}
    \caption{Example speculative-parallel execution over iterations 0--3. Here, $s_t$ denotes the amount of speculative execution at iteration $t$; the superscripts $a$ and $r$ denote accepted and rejected speculation, respectively; and $n_t$ denotes the amount of normal execution at $t$. Below the timeline, we illustrate how tokens accepted at iteration $i$ are used for speculative execution in iteration $i+1$ ($s_i^a\rightarrow s_{i+1}$), while rejected tokens are re-executed in iteration $i+1$ ($s_i^r\rightarrow n_{i+1}$, denoted by red arrow). 
    }
    \label{fig:Recurrence-Relation}
\end{figure}

\noindent
{\bf Cost model for PP, TP, and SP.} 
We develop a cost model for the steady-state decoding throughput of TP, PP, and SP, each using $N$ GPUs. PP and SP both use $N$ pipeline stages. For simplicity, we assume all sequences have equal length. 

Let $B$ denote the maximum number of sequences that can be processed in a batch under TP. For PP, assuming $N$ pipeline stages with $N$ micro-batches, the effective batch size per stage becomes $B/N$. For SP, we estimate its effective batch size, defined as the expected number of accepted tokens in a batch, using the recurrence relation shown below and illustrated by the example in Figure~\ref{fig:Recurrence-Relation}.
{\setlength{\jot}{2pt}
\begin{equation}
\begin{aligned}
s^a_t &= (s^a_{t-1}+n_{t-1})\Theta_{t-1},\\
s^r_t &= \sum_{i=1}^{N-1}(s^a_{t-i}+n_{t-i})(1-\Theta_{t-i}),\\
n_t &= (s^a_{t-N}+n_{t-N})(1-\Theta_{t-N}),\\
\end{aligned}
\label{eq:sp_flow}
\end{equation}
}

Here, $s^*_t$ and $n_t$ denote the numbers of sequences processed by speculative execution and normal execution, respectively, at iteration $t$, where $s^*_t + n_t = B$. Similarly, $s^a_t$ and $s^r_t$ denote the numbers of accepted and rejected speculative executions, respectively. Let $\Theta_t$ denote the speculation accuracy at iteration $t$. For simplicity, we assume that the speculation accuracy is time-invariant, i.e., $\Theta_t = \Theta$. The effective batch size at iteration $t$ is therefore $B_{\mathrm{eff}}(t) = s^a_t + n_t$, and the steady-state effective batch size $B_{\mathrm{eff}}^{\star}$ is given as follows.
\begin{equation}
B_{\mathrm{eff}}^{\star}
= \lim_{t\to\infty} B_{\mathrm{eff}}(t)
= \frac{B}{N-\Theta(N-1)}.
\label{eq:sp_beff}
\end{equation}

\begin{table}[t]
\centering
\footnotesize
\setlength{\tabcolsep}{5pt}
\renewcommand{\arraystretch}{1.1}
\resizebox{\columnwidth}{!}{
\begin{tabular}{c c c c}
\hline
{\bf PP} & {\bf TP} & {\bf SP} & {\bf SP$_{\alpha}$} \\
\hline
$\dfrac{B/N}{T_{\mathrm{pp}}(B/N)}$ 
& $\dfrac{B}{T_{\mathrm{tp}}(B)}$ 
& $\dfrac{B_{\mathrm{eff}}^{\star}}{T_{\mathrm{sp}}(B)}$ 
& $\dfrac{B_{\mathrm{eff,\alpha}}^{\star}}{T_{\mathrm{sp}}(B_{\mathrm{\alpha}}^{\star})}$ \\
\hline
\end{tabular}
}
\caption{Estimation of throughput for PP, TP, SP and SP$_{\alpha}$. 
$N$ denotes the number of pipeline stages, 
$B$ the maximum batch size, and $T_*(\cdot)$ the single-stage decoding time for a given batch size. SP$_{\alpha}$ incorporates selective speculation described in Section~\ref{sec:alpha-speculation}.}
\label{tab:throughput}
\end{table}

We next estimate the decoding throughput of PP, TP, and SP using their effective batch sizes, which correspond to the numbers of decoded tokens, together with their execution times. To this end, we obtain the decoding-latency functions $T_{\mathrm{pp}}(\cdot)$, $T_{\mathrm{tp}}(\cdot)$, and $T_{\mathrm{sp}}(\cdot)$ for PP, TP, and SP, respectively, through profiling. Table~\ref{tab:throughput} summarizes the resulting throughput estimates, which we use to determine the optimal combination of PP, TP, and SP. 

\begin{figure}[t]
   \centering
   \includegraphics[width=1.0\linewidth]{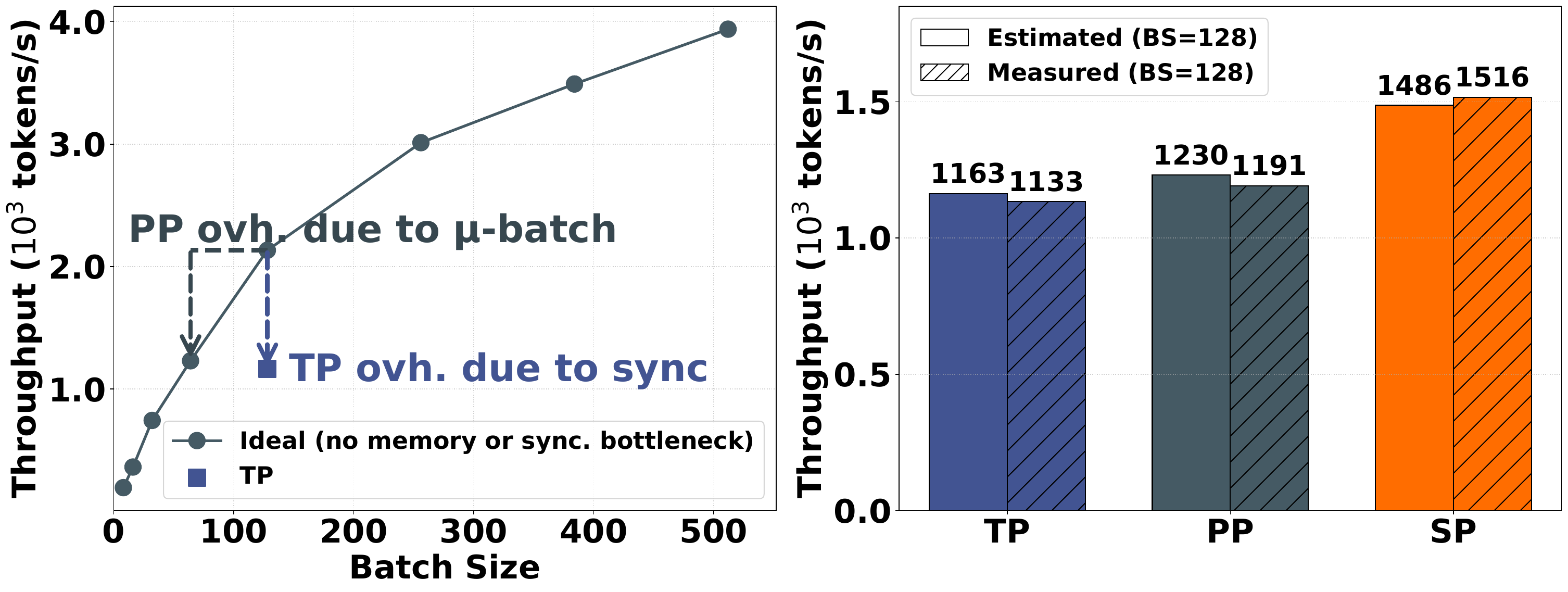}
   \caption{Throughput overheads of PP and TP (left), and comparison of the throughput estimated by our cost model (Table~\ref{tab:throughput}) and the measured throughput (right).}
   \label{fig:cost-model-confirm}
\end{figure}

The estimation captures the memory overhead of PP from micro-batching and the synchronization overhead of TP, as illustrated in Figure~\ref{fig:cost-model-confirm} (left). The right plot compares estimated and measured throughput. The estimates are highly accurate, with only small errors, mostly due to batch-size variation, supporting its use for selecting the optimal parallel configuration.

\section{Lookahead Scheduling Optimization} 
\label{sec:lookahead}

We further improve the efficiency of speculative-parallel execution using two prediction-based techniques. The first predicts the acceptance probability of draft tokens and selects which tokens to speculate on to maximize throughput. The second predicts the remaining output length of each sequence and uses this information to delay the scheduling of incoming requests when necessary, thereby minimizing evictions of running requests.

\subsection{Acceptance-Guided Selective Speculation}
\label{sec:alpha-speculation}

In self-speculative models, the confidence of a draft token is often used to estimate its acceptance probability~\cite{kangaroo, eagle2}. Spexis also uses this confidence, but rather than explicitly predicting acceptance, it ranks drafted tokens by confidence, discards low-ranked tokens, and speculates only on high-ranked ones.

More specifically, Spexis drops a fixed $\alpha$ fraction of drafted tokens in a batch and performs speculation on the remaining tokens. Our analysis shows that a small fraction of low-confidence draft tokens in each batch has little chance of being accepted; thus, dropping them largely improves speculation accuracy. To determine $\alpha$, we profile the acceptance rate of the retained draft tokens. By jointly considering this acceptance rate and the latency of speculation, we identify the optimal $\alpha$. In our evaluation, we used an $\alpha$ value in the range of 0.15--0.3.

We extend the cost model in the previous section to reflect selective speculation. Let $\alpha \in [0,1)$ denote the drop ratio, $\Theta_{\alpha}$ the speculation accuracy under $\alpha$ (with $\Theta_{0}=\Theta$), and $B_t$ the total batch size at iteration $t$, including both speculative and normal executions. The steady-state batch size $B_{\alpha}^{\star}$ is then computed by the recurrence relation below (for simplicity, and since $\alpha$ is small, we assume that dropping affects only speculative execution):

\begin{equation}
\begin{aligned}
B_{t+1} &= \alpha B_{t+1-N} + (1-\alpha)B_t, \\
B_{\alpha}^{\star}
&= \lim_{t\to\infty} B_t
 = \frac{B}{1+\alpha (N-1)} ,
\end{aligned}
\label{eq:scheme51_run}
\end{equation}

\noindent 
The expected effective batch size, counting only accepted tokens from speculative and normal executions, is then computed using Equations~\ref{eq:sp_beff} and~\ref{eq:scheme51_run}:
\begin{equation}
B_{\mathrm{eff},\alpha}^{\star}
=
\alpha B_{\alpha}^{\star}
+
\frac{(1-\alpha)B_{\alpha}^{\star}}
     {N-\Theta_\alpha(N-1)} .
\label{eq:scheme51_beff}
\end{equation}

\noindent 
Finally, the throughput of SP with selective speculation is computed as the effective batch size divided by the latency of processing the total batch size, i.e., $\frac{B_{\mathrm{eff},\alpha}^{\star}} {T_{\mathrm{sp}}(B_{\alpha}^{\star})}$ as shown in Table~\ref{tab:throughput} under SP${_\alpha}$.

\subsection{Remaining-Length-Aware Scheduling}
\label{sec:length-scheduling}

SP is memory-efficient because speculative execution shares its KV cache with the normal execution. It further reduces memory pressure by predicting the remaining lengths of running sequences and using this information to delay the scheduling of new requests. Specifically, we estimate the approximate remaining lengths of running sequences to predict their KV cache usage in upcoming decoding iterations. Based on this estimate, we determine when to schedule new requests to prevent eviction of running sequences and their KV cache entries.

For remaining-length prediction, rather than predicting the exact length, we use ordinal-threshold prediction to estimate whether the remaining length falls below 4, 8, 16, 32, and 64 tokens, or exceeds 64 tokens. To make this prediction, we train a three-layer MLP that takes the hidden state of the draft layer as input and outputs five confidence scores corresponding to the five thresholds (4--64). For each threshold, we choose a confidence cutoff such that, when the model predicts that the remaining length falls below that threshold, the prediction attains high precision (83\%, as shown in our evaluation) while maintaining reasonable recall. 

With the predicted remaining lengths, we estimate KV cache memory usage over the next 64 decoding iterations and determine whether admitting a new request during that window would require eviction (Algorithm~1 describes this estimation process in detail). We jointly consider the throughput benefit of admitting the new request earlier and the recomputation cost incurred by evicting running sequences. Although the recomputation penalty is often larger in practice, the benefit of admitting a new request earlier may sometimes outweigh the penalty. Hence we explicitly model this trade-off.

\providecommand{\Notation}{\item[\textbf{Notation:}]}
\providecommand{\CommentFont}[1]{\textit{#1}}
\providecommand{\LComment}[1]{\Statex \CommentFont{$\triangleright$ #1}}
\algrenewcommand{\algorithmiccomment}[1]{\hfill\CommentFont{$\triangleright$ #1}}
\makeatletter
\providecommand{\algrule}[1][0.4pt]{%
  \Statex\hspace*{-\dimexpr\ALG@thistlm+\leftmargin\relax}%
  \rule{\dimexpr\linewidth+\ALG@thistlm+\leftmargin\relax}{#1}}
\makeatother

\begin{algorithm}[t]
  \caption{Look-ahead memory estimation over future iterations}
  \label{alg:memory-est-combined}
  \footnotesize
  \begin{algorithmic}[1]
    \renewcommand{\algorithmicrequire}{\textbf{Input:}}
    \renewcommand{\algorithmicensure}{\textbf{Output:}}
    \Require $\mathcal{B}$: set of running requests $r$
    \Statex $pp$: pipeline depth; \ $\theta$: speculation accuracy
    \Statex $\mathcal{E}(\cdot)$: length prediction precision
    \Statex $\ell^{(r)},\,\tau^{(r)}$: current and remaining length of request $r$
    \Ensure $\mathcal{M}$: list of estimated memory usage (in tokens) in next 1..64 iterations
    \Notation Per request, let $w$ count speculation rejections;
    \Statex $t_{i,w} \triangleq i - pp\,w$ \hfill \CommentFont{$\triangleright$ \# accepted tokens at iteration $i$}
    \Statex $k_{i,w} \triangleq t_{i,w} + w$ \hfill \CommentFont{$\triangleright$ \# generated tokens  at iteration $i$}
    \Statex $p_{i,w} \triangleq \binom{k_{i,w}}{t_{i,w}}\,\theta^{t_{i,w}}(1-\theta)^{w}$ \hfill \CommentFont{$\triangleright$ $\Pr[\#\text{generated} = k_{i,w}]$}
    \Statex $\mathbf{1}[\cdot]$: indicator function \hfill \CommentFont{$\triangleright$ $1$ if true, $0$ otherwise}
    \algrule
    \State $\mathcal{M} \gets [\,]$, \ $L_{\mathrm{total}} \gets \displaystyle\sum_{r \in \mathcal{B}} \ell^{(r)}$
    \ForAll{$i \in \{1,\dots,64\}$}
      \State $w_{\max} \gets \lfloor i/pp \rfloor$
      \LComment{expected \#generated tokens per request}
      \State $\mathbb{E}[L_{\mathrm{gen}}] \gets \displaystyle\sum_{w=0}^{w_{\max}} k_{i,w}\, p_{i,w}$
      \LComment{completion probability of each request $r \in \mathcal{B}$}
      \State $P_{\mathrm{fin}}^{(r)} \gets \mathcal{E}(\tau^{(r)})\,\displaystyle\sum_{w=0}^{w_{\max}} \mathbf{1}\bigl[k_{i,w} \ge \tau^{(r)}\bigr]\, p_{i,w}$
      \State $\mathbb{E}[N_{\mathrm{fin}}] \gets \displaystyle\sum_{r \in \mathcal{B}} P_{\mathrm{fin}}^{(r)}$ \Comment{expected \#finished requests}
      \State $\mathbb{E}[L_{\mathrm{fin}}] \gets \displaystyle\sum_{r \in \mathcal{B}} \ell^{(r)}\, P_{\mathrm{fin}}^{(r)}$ \Comment{memory they release}
      \LComment{memory estimate at look-ahead iteration $i$}
      \State $M_{\mathrm{est}} \gets L_{\mathrm{total}} + \bigl(|\mathcal{B}| - \mathbb{E}[N_{\mathrm{fin}}]\bigr)\,\mathbb{E}[L_{\mathrm{gen}}] - \mathbb{E}[L_{\mathrm{fin}}]$
      \State Append $M_{\mathrm{est}}$ to $\mathcal{M}$
    \EndFor
    \State \Return $\mathcal{M}$
  \end{algorithmic}
\end{algorithm}

\noindent
\textbf{Analytic model for length-aware scheduling.}
We estimate the throughput over the period between two request arrivals.
Let $I$ denote the number of iterations in this period, $bs$ the number of running decode requests before the new arrival, and $t_d$ and $t_p$ the latency of one decode iteration and one prefill, respectively. Then the prefill-to-decode (or recompute-to-decode) ratio is $pdr := t_p/t_d$. 

We compare two scheduling policies: \emph{aggr} (aggressive), which runs the new request immediately and may later incur eviction and recomputation, and \emph{delay} (ours), which runs the request at the earliest expected iteration that does not trigger eviction.

We consider the case where $n$(>$0$) evictions occur under \emph{aggr} during the period. Let $i_{\mathrm{evict}}^k$ and $i_{\mathrm{rec}}^k$ denote the eviction iteration and the subsequent recomputation iteration for the $k$-th eviction, respectively. We define $i_{\mathrm{wait}}^k := i_{\mathrm{rec}}^k - i_{\mathrm{evict}}^k$ as the number of iterations the $k$-th evicted sequence remains unscheduled before recomputation. Let $i_{\mathrm{delay}}$ denote the earliest scheduling iteration under \emph{delay} that avoids eviction. The throughput ratio, or gain, $G$, of \emph{delay} over \emph{aggr} is then computed as follows.
\begin{equation}
\nonumber
G =
\frac{I + (n+1)(pdr-1)}{I + pdr - 1} \cdot
\frac{bs+1 - \dfrac{i_{\mathrm{delay}}}{I}}
     {bs+1 - \sum\limits_{k=1}^n \dfrac{i_{\mathrm{wait}}^k}{I}}
\label{eq:delay_gain}
\end{equation}

The details are provided in the appendix. 
The left term of the product captures the recomputation overhead, while the right term captures the cost of delaying scheduling in \emph{delay} relative to waiting in \emph{aggr}. Since $bs$ is typically much larger than the subtracted terms, the right term is usually close to one, making the recomputation overhead the main factor in the gain of \emph{delay}. However, if evictions are rare (e.g., only one) and $i_{\mathrm{delay}} \gg i_{\mathrm{wait}}$, the gain can be less than one. We account for both effects when scheduling new requests.

\section{Implementation}
\label{sec:implementation}
We implemented our prototype on vLLM (v0.8.4) and Ray (v2.53.0) by extending \texttt{GPURunner} to add speculative-parallel execution. Our implementation remains compatible with existing optimizations in vLLM and Ray, particularly Ray's compiled DAG optimization, which reduces RPC overhead. However, because the current DAG implementation exposes stage outputs only after the full model execution completes, we use a separate IPC mechanism to obtain the first-stage output for drafting.

Spexis places an LM head at the first and last pipeline stages. Because the additional LM head can imbalance the pipeline stages, we offload the input embedding, whose parameter size is comparable to that of the LM head, to host memory and perform the lookup on the CPU. This adds only a few hundred microseconds of overhead and has little impact on overall performance.

For the drafting adapter, we follow prior work~\cite{kangaroo} and design it with a linear layer and an attention layer, followed by the (pre-trained) LM head. We trained the adapter on the ShareGPT dataset using the AdamW optimizer for five epochs. For Llama-3.3-70B, this training took about 12 hours on a single H200 GPU.

For the length prediction model, we use a three-layer MLP with two GELU activations. We train the model on ShareGPT to predict the confidence that the remaining length is below 4, 8, ..., and 64 tokens. For training, we first generate the full output sequence from each input and then use the generated output to derive the remaining lengths. We train the model for 20 epochs, which took about 15 hours on a single H100 GPU.

\begin{figure*}[t]
    \centering
    \includegraphics[width=\textwidth]{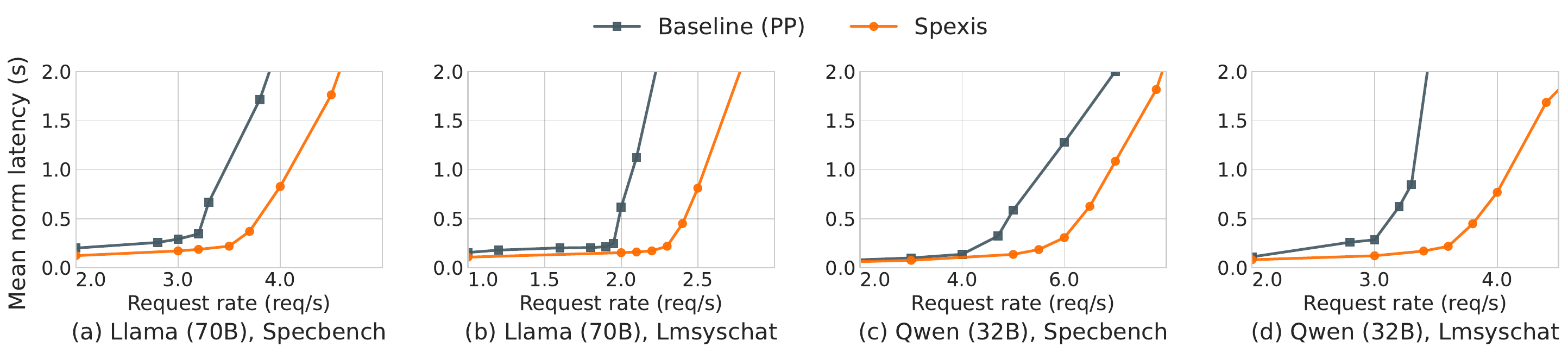}
    \caption{Mean normalized latency of Spexis and PP as the request rate increases in the inter-node setting.}
    \label{fig:intrernode}
\end{figure*}

\begin{figure*}[t]
    \centering
    \includegraphics[width=\textwidth]{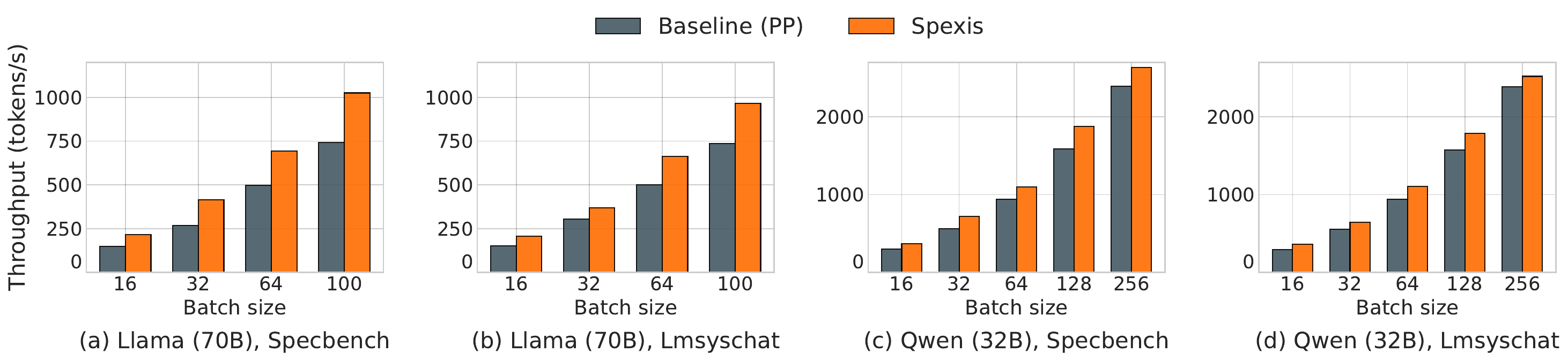}
    \caption{Decoding throughput of Spexis and PP as the batch size increases in the inter-node setting.}
    \label{fig:Decoding-Throughput}
\end{figure*}

\section{Evaluation}
\label{sec:eval}

We evaluate Spexis against PP and TP, including their optimal combinations. 
We conduct experiments using resources from both our private cluster and a public cloud platform. Across these environments, we evaluate two widely adopted model families, Llama 3.3~\cite{meta2024llama33} and Qwen3~\cite{qwen3technicalreport}. We use SpecBench~\cite{xia-etal-2024-unlocking}, LMSYS-Chat~\cite{zheng2023lmsyschat1m} and UltraChat~\cite{ding2023enhancing} as evaluation datasets.

Our private cluster provides two configurations: (1) an inter-node configuration with two nodes, each equipped with one NVIDIA RTX PRO 6000 GPU and connected via 1 Gb Ethernet; and (2) an intra-node configuration with a single node equipped with eight NVIDIA A40 GPUs communicating over PCIe 4.0$\times$16. We use these configurations to measure the overall performance gains of Spexis and provide detailed breakdowns.

To cover a broader range of GPU and interconnect configurations, we use four hardware configurations on Runpod~\cite{runpod2026cloudgpus}, a GPU cloud provider offering access to a diverse range of accelerator instances. Each configuration consists of a single eight-GPU node: (1) H100 SXM with an NVSwitch-based NVLink fabric; (2) A100 SXM with an NVSwitch-based NVLink fabric; (3) RTX PRO 6000 with PCIe 5.0 $\times$16; and (4) L40S with PCIe 4.0 $\times$16. On these instances, we compare TP and SP to assess the effectiveness of SP on standard single-node, multi-GPU servers across diverse GPU and interconnect configurations.

\subsection{Overall Performance Improvements}

We evaluated the speedup of Spexis in both inter-node and intra-node settings on our private cluster. 
To evaluate end-to-end serving performance, we conducted a standard load-scaling experiment. Specifically, for each request rate, we submitted requests from SpecBench or LMSYS-Chat to the serving engine, with request arrivals following a Poisson process. We varied the request rate across runs and measured the latency metrics for each run. As the latency metric, we report mean normalized latency, i.e., the average end-to-end latency of each request normalized by its output length. This metric is widely used for evaluating LLM inference performance~\cite{orca, vllm}.

We first report the results in the inter-node setting. Figure~\ref{fig:intrernode} shows the mean normalized latency of Llama and Qwen on the SpecBench and LMSYS-Chat datasets. The baseline uses a two-stage pipeline with two micro-batches. Spexis also uses a two-stage pipeline, with all optimizations in Section~\ref{sec:lookahead} enabled. We use a drop rate $\alpha$ of 0.15 for Llama and 0.3 for Qwen. We see from the figure that Spexis can serve up to 34\% higher request rates than the baseline. The largest gain is observed for Qwen on LMSYS-Chat, where Spexis serves 34\% more requests per second. 

\begin{figure}[t]
    \centering
    \includegraphics[width=\columnwidth]{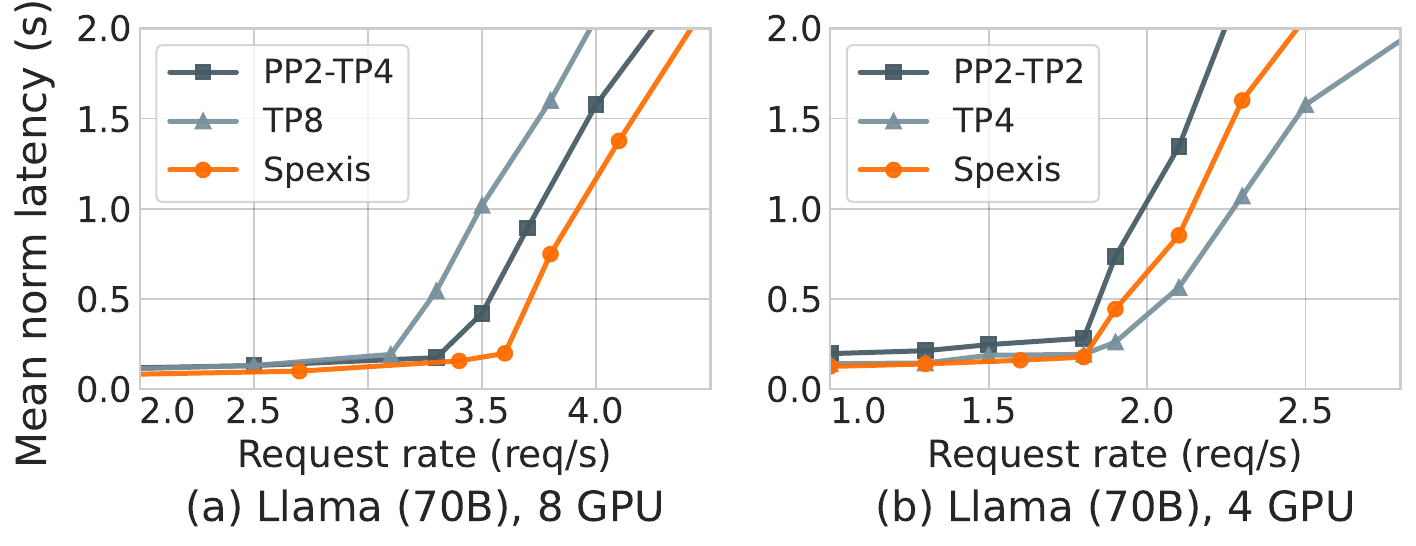}
    \caption{Mean normalized latency of Spexis, TP, and PP, including their best TP+PP configuration, as the request rate increases in the intra-node setting.}
    \label{fig:Mean-Normalized-Latency}
\end{figure}

We report decoding performance separately in Figure~\ref{fig:Decoding-Throughput}. Under disaggregation, prefill and decoding run on separate devices, making overall performance largely dependent on decoding throughput. We compare the decoding throughput of the baseline and Spexis as the batch size increases from 16 to 256. Spexis consistently outperforms the baseline across all batch sizes, with larger gains at higher batch sizes. At batch sizes of 64--100 for Llama and 128--256 for Qwen, Spexis achieves up to 54\% higher throughput.

Now we report the serving performance for the intra-node setting. Figure~\ref{fig:Mean-Normalized-Latency} compares the normalized latency of Spexis with those of PP and TP, including their optimal combination, for Llama-3.3-70B. The left plot shows the results on eight GPUs, where Spexis outperforms the other two settings by 8\% and 16\%. On four GPUs, Spexis outperforms the PP+TP configuration with both degrees set to two, but remains slower than the TP-only setting. The cost model in Table~\ref{tab:throughput} explains this small gap. For TP, communication latency is captured by the $T_{\mathrm{TP}}$ term, which is relatively small in the four-GPU configuration without inter-socket communication; using the profiled value of $T_{\mathrm{TP}}$, the cost model estimates the throughput overhead as 21\%. For SP, the speculation accuracy obtained from offline profiling determines $B_{\mathrm{eff}}^{\star}$, which translates into a 21.5\% throughput overhead. This small difference in modeled overhead is consistent with the slight measured advantage of TP-4 over SP. Overall, these results show that, in intra-node environments, Spexis is particularly effective for larger models that require eight or more GPUs.

\subsection{Performance Breakdown}

We examine the effects of the two components of the lookahead scheduling optimization described in Section~\ref{sec:lookahead}: selective speculation and remaining-length-aware scheduling. For this breakdown, we measure performance as we incrementally add speculative-parallel execution, selective speculation, and remaining-length-aware scheduling to the baseline, using Llama-3.3-70B and Qwen3-32B on SpecBench.

\begin{figure}[!ht]
    \centering
    \includegraphics[width=\columnwidth]{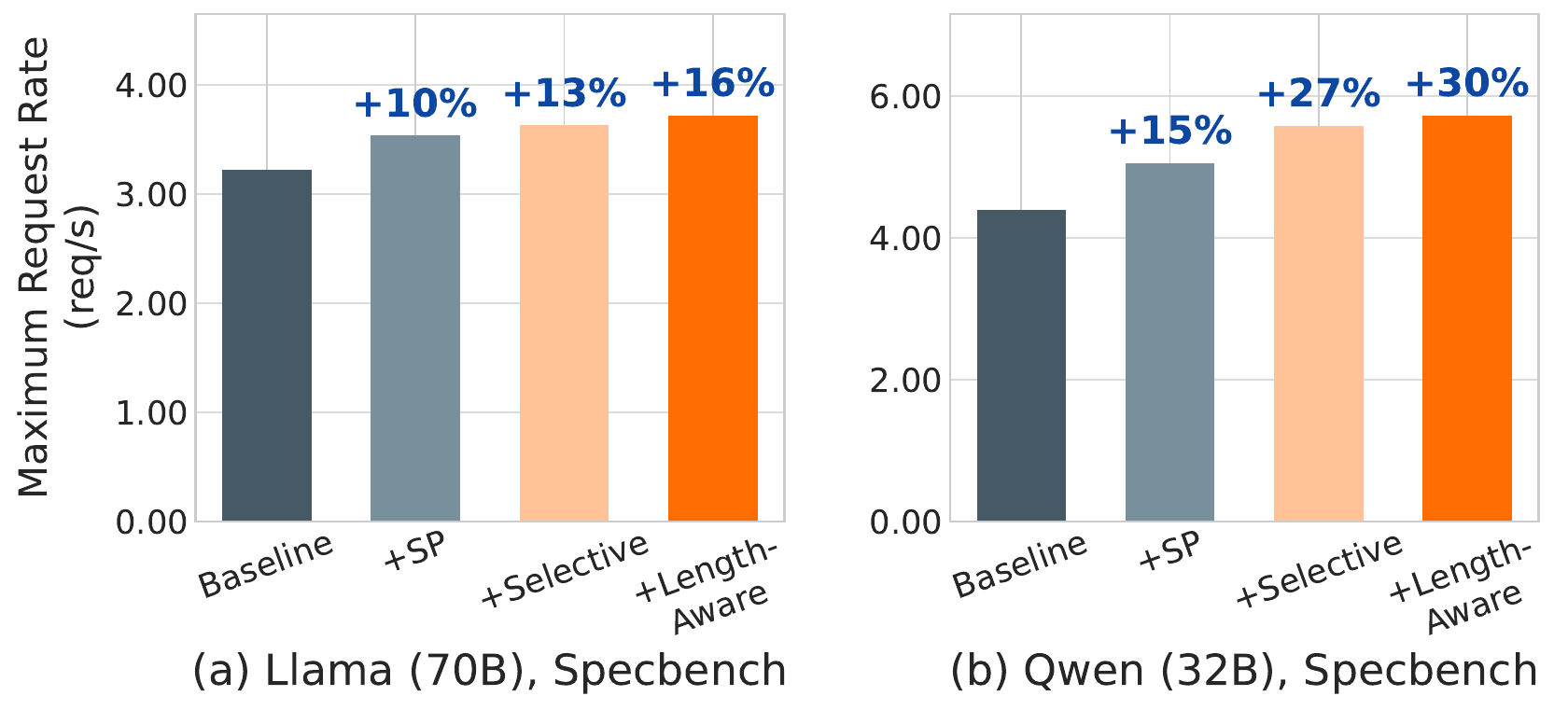}
    \caption{Performance breakdown of Spexis on SpecBench. From left to right, we incrementally add speculative-parallel execution (SP), selective speculation (Selective), and remaining-length-aware scheduling (Length-Aware), with percentages showing cumulative improvements over the baseline.}
    \label{fig:perf-breakdown}
\end{figure}

Figure~\ref{fig:perf-breakdown} shows the resulting performance breakdown. Speculative-parallel execution alone improves performance by 10--15\%. Selective speculation provides additional gains of 3~pp and 12~pp for Llama and Qwen, respectively, by dropping the $\alpha$ fraction of low-confidence draft tokens. It has a larger impact on Qwen, as the model has lower speculation accuracy. Remaining-length-aware scheduling provides a further 3~pp gain for both models by delaying admissions that would otherwise evict running requests and require KV-cache recomputation.

\subsection{Performance on Cloud GPUs}

We report experiments on cloud GPU instances to assess the effectiveness of Spexis across diverse intra-node GPU and interconnect configurations. Based on a survey of representative single-node, eight-GPU offerings from major cloud providers, we select four Runpod configurations---H100 SXM, A100 SXM, RTX PRO 6000, and L40S---covering both NVSwitch-based NVLink fabrics and PCIe interconnects. For each configuration, we compare Spexis against TP-only using Llama-3.3-70B and Qwen3-32B on SpecBench. Due to resource constraints, we report fixed-batch decoding throughput rather than the serving performance measured on our private cluster.

Table~\ref{tab:cloud-throughput} reports the results. Spexis outperforms TP-only in five of the eight model--hardware pairs. In particular, Spexis is faster in all four PCIe cases, achieving speedups of 1.02--1.21$\times$, and achieves a 1.27$\times$ speedup for Llama-3.3-70B on A100 SXM. TP remains faster in the other three SXM cases because its communication overhead is lower than the speculation-failure overhead incurred by Spexis. Overall, these results demonstrate the effectiveness of Spexis across a range of intra-node GPU and interconnect configurations: Spexis consistently outperforms TP on the evaluated PCIe systems and can also outperform TP in a certain high-bandwidth SXM configuration. 

\begin{table}[t]
\centering
\footnotesize
\setlength{\tabcolsep}{2pt}
\begin{tabular*}{\columnwidth}{@{\extracolsep{\fill}} llcrrr @{}}
\toprule
\multirow{2}{*}{GPU} &
\multirow{2}{*}{Interconnect} &
\multirow{2}{*}{Model} &
\multicolumn{2}{c}{Thpt. (tokens/s)} &
\multirow{2}{*}{Speedup} \\
\cmidrule(lr){4-5}
 & & & Spexis & TP & \\
\midrule

\multirow{2}{*}{\makecell[l]{H100\\SXM}}
 & \multirow{2}{*}{\makecell[l]{NVLink via\\NVSwitch}}
 & L70 & 3587 & \textbf{3652} & 0.98$\times$ \\
 & & Q32 & 2402 & \textbf{5068} & 0.47$\times$ \\

\midrule

\multirow{2}{*}{\makecell[l]{A100\\SXM}}
 & \multirow{2}{*}{\makecell[l]{NVLink via\\NVSwitch}}
 & L70 & \textbf{1228} & 966 & 1.27$\times$ \\
 & & Q32 & 1454 & \textbf{2058} & 0.71$\times$ \\

\midrule

\multirow{2}{*}{\makecell[l]{RTX PRO\\6000}}
 & \multirow{2}{*}{\makecell[l]{PCIe 5.0\\$\times$16}}
 & L70 & \textbf{1492} & 1236 & 1.21$\times$ \\
 & & Q32 & \textbf{2201} & 2062 & 1.07$\times$ \\

\midrule

\multirow{2}{*}{L40S}
 & \multirow{2}{*}{\makecell[l]{PCIe 4.0\\$\times$16}}
 & L70 & \textbf{993} & 930 & 1.07$\times$ \\
 & & Q32 & \textbf{1509} & 1473 & 1.02$\times$ \\

\bottomrule
\end{tabular*}

\caption{Fixed-batch decoding throughput (tokens/s) of Spexis and TP-only
on single-node, eight-GPU Runpod configurations, measured at a fixed batch size of 64.
Speedup is computed as Spexis throughput divided by TP throughput;
bold indicates the higher throughput.}
\label{tab:cloud-throughput}
\end{table}

\subsection{Accuracy of Drafting Adapter}

We evaluate the speculation accuracy of the self-speculative models used in our experiments. We fine-tune the models following prior work, while inserting the drafting adapter at the 1/2 point of the model for two-stage pipeline execution and at the 1/4 point for four-stage pipeline execution.

Table~\ref{tab:accuracy_compact} reports the speculation accuracy across all SpecBench subtasks for Llama-70B and Qwen-32B. When the first half of the model is used for drafting, the accuracy is moderately high, ranging from 0.60 to 0.79. When only the first quarter is used, the accuracy decreases to 0.51--0.63.

\begin{figure}[h]
    \centering
    \includegraphics[width=\columnwidth]{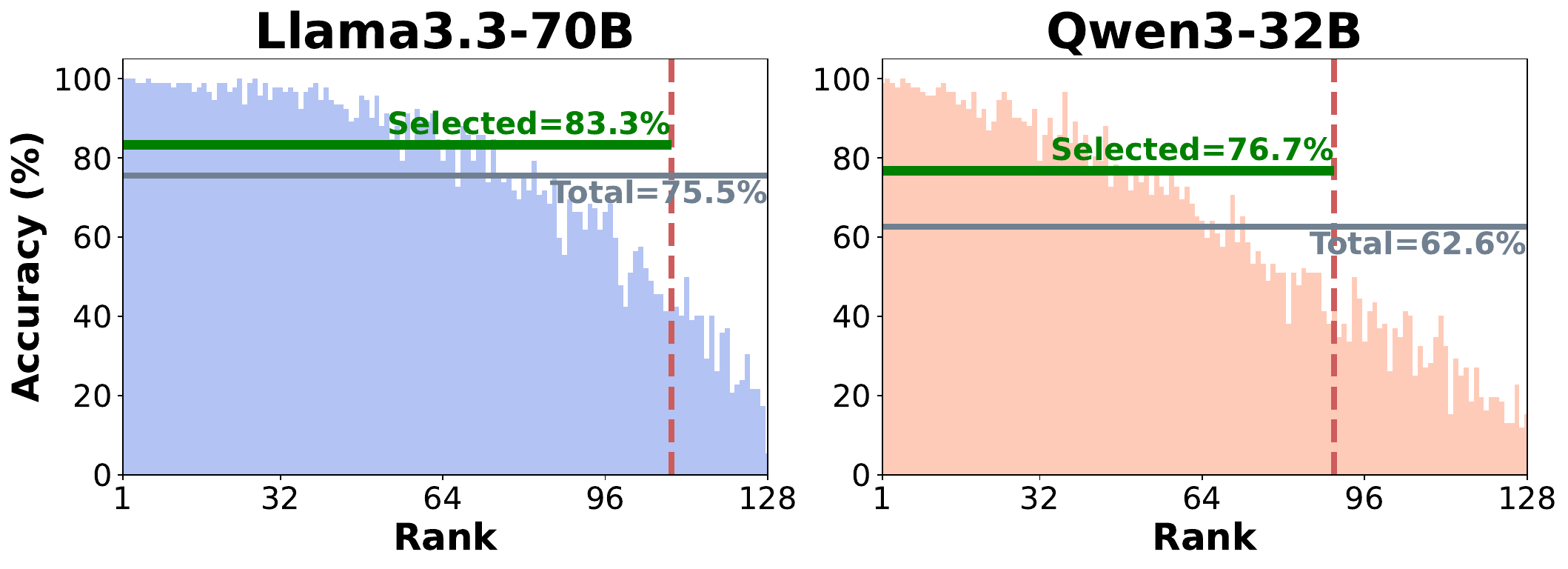}
    \caption{
    Speculation accuracy ranked by confidence score on SpecBench Dataset (both 1/2 exited). The value of cutoff $\alpha$ is specified by dotted vertical lines.
    }
    \label{fig:acc-by-rank}
\end{figure}

Spexis further improves speculation accuracy by selectively applying speculation to a subset of sequences, as described in Section~\ref{sec:alpha-speculation}. Specifically, we rank the sequences in a batch by their speculation confidence and drop the lowest $\alpha$ fraction, speculating only on the higher-confidence ones. Figure~\ref{fig:acc-by-rank} shows the acceptance ratio of tokens as a function of their confidence rank, along with the $\alpha$ values selected for the two models. The results show that dropping the lowest-confidence $\alpha$ fraction and speculating only on the higher-ranked sequences improves speculation accuracy by 8--14\%.

\begin{table}[ht]
\centering
\footnotesize
\setlength{\tabcolsep}{3pt} 
\begin{tabular*}{\columnwidth}{@{\extracolsep{\fill}} llccccccc @{}}
\toprule
\multirow{2}{*}{Model} & \multirow{2}{*}{\makecell{Exit}} & \multicolumn{6}{c}{Task Accuracy} & \multirow{2}{*}{Avg.} \\
\cmidrule(lr){3-8}
 &  & Math & QA & Sum & RAG & Tran & Mult &  \\
\midrule
\multirow{2}{*}{L70} & 1/2 & .840 & .743 & .760 & .788 & .774 & .793 & \textbf{.785} \\
                     & 1/4 & .680 & .629 & .581 & .622 & .570 & .656 & \textbf{.634} \\
\midrule
\multirow{2}{*}{L8}  & 1/2 & .800 & .776 & .660 & .727 & .644 & .759 & \textbf{.759} \\
                     & 1/4 & .665 & .609 & .487 & .547 & .461 & .625 & \textbf{.596} \\
\midrule
\multirow{2}{*}{Q32} & 1/2 & .682 & .535 & .565 & .619 & .609 & .603 & \textbf{.596} \\
                     & 1/4 & .597 & .471 & .474 & .489 & .427 & .550 & \textbf{.511} \\
\midrule
\multirow{2}{*}{Q8}  & 1/2 & .674 & .559 & .598 & .608 & .586 & .620 & \textbf{.637} \\
                     & 1/4 & .601 & .494 & .515 & .521 & .469 & .557 & \textbf{.532} \\
\bottomrule

\end{tabular*}

\caption{Speculation accuracy on SpecBench for Llama-3.3-70B-Instruct (L70), Llama-3.1-8B-Instruct (L8), Qwen3-32B (Q32), and Qwen3-8B (Q8). The drafting adapter (Exit) is inserted at either the halfway point (1/2) or the one-quarter point (1/4) of the model.}
\label{tab:accuracy_compact}
\end{table}

\begin{table}[t]
\centering
\footnotesize
\setlength{\tabcolsep}{3pt}
\renewcommand{\arraystretch}{1.05}
\resizebox{\columnwidth}{!}{%
\begin{tabular}{l l cc cc cc} 
\toprule
& & \multicolumn{2}{c}{ShareGPT} & \multicolumn{2}{c}{SpecBench} & \multicolumn{2}{c}{UltraChat} \\
\cmidrule(lr){3-4} \cmidrule(lr){5-6} \cmidrule(lr){7-8}
Model & Thresh. & Prec. & Recall & Prec. & Recall & Prec. & Recall \\
\midrule

\multirow{5}{*}{L70}
& $L < 4$   & 79.0\% & 53.4\% & 82.6\% & 59.7\% & 76.3\% & 56.0\% \\
& $L < 8$   & 78.6\% & 55.2\% & 80.3\% & 59.5\% & 78.1\% & 57.6\% \\
& $L < 16$  & 81.2\% & 58.1\% & 85.8\% & 59.4\% & 81.5\% & 61.9\% \\
& $L < 32$  & 83.5\% & 64.0\% & 88.9\% & 61.3\% & 82.9\% & 68.6\% \\
& $L < 64$  & 85.2\% & 67.6\% & 90.7\% & 67.5\% & 85.8\% & 71.1\% \\
\midrule

\multirow{5}{*}{Q32}
& $L < 4$   & 76.1\% & 53.9\% & 70.0\% & 39.8\% & 63.5\% & 52.2\% \\
& $L < 8$   & 78.4\% & 58.8\% & 73.9\% & 43.7\% & 67.5\% & 54.5\% \\
& $L < 16$  & 82.6\% & 65.8\% & 77.6\% & 52.4\% & 73.0\% & 61.5\% \\
& $L < 32$  & 87.0\% & 69.0\% & 83.5\% & 60.8\% & 80.1\% & 69.3\% \\
& $L < 64$  & 85.7\% & 62.8\% & 83.1\% & 67.7\% & 82.3\% & 71.0\% \\
\bottomrule
\end{tabular}%
}
\caption{Precision (Prec.) and recall for the remaining-length prediction using thresholds of 4, ..., 64 tokens (L70: Llama-3.3-70B-Instruct, Q32: Qwen3-32B).} 
\label{tab:length-prec-recall}
\end{table}

\subsection{Accuracy of Length Prediction}

We evaluate the accuracy of our remaining-length prediction method. For the prediction, we trained a three-layer MLP model that outputs confidence scores for each sequence if its remaining length falls below 4, 8, 16, 32, and 64 tokens. Then we choose a confidence cutoff for each threshold (4, 8, ..., 64) based on profiling to predict if the remaining length is below the corresponding threshold.

Table~\ref{tab:length-prec-recall} shows the precision and recall of the prediction method.
The MLP is trained on a subset of ShareGPT, and the reported scores are measured on the remaining ShareGPT data as well as the full SpecBench and UltraChat datasets. The precision is generally high across datasets and length thresholds, mostly ranging from 70\% to 90.7\%. The recall is also moderately high ranging 40--71\%.

\section{Conclusion}
\label{sec:conclusion}
We present Spexis, a multi-GPU scheduling framework that improves the efficiency of LLM inference by integrating speculative decoding with distributed execution. Spexis introduces speculative parallelism, which overlaps speculative and normal execution to expose an additional axis of parallelism without increasing KV-cache memory usage. It further incorporates lookahead scheduling to make admission and execution decisions based on predicted speculation quality and near-future memory pressure, thereby reducing wasted speculation, KV-cache eviction, and recomputation.

Our evaluation shows that Spexis consistently improves serving performance across diverse GPU configurations, achieving up to 34\% speedups over the baseline using the optimal combination of pipeline and tensor parallel execution. The results show that speculative decoding can be leveraged not only as an algorithmic acceleration technique, but also as an effective systems mechanism for improving the performance of distributed LLM inference.

\section*{Limitations}
Our exploration of Spexis leaves several areas for future investigation. First, the framework requires training auxiliary components—specifically, a drafting adapter and a length predictor. While this preliminary overhead is relatively small, it introduces an extra step in the deployment pipeline. Second, the performance of Spexis is inherently sensitive to speculation accuracy. Although we mitigate the impact of low-accuracy scenarios using the fallback mechanisms described in Section~\ref{sec:alpha-speculation}, significant domain shifts may still reduce the expected throughput gains. Finally, the relative advantage of Speculative Parallelism (SP) over Tensor Parallelism (TP) is highly dependent on hardware interconnects. In environments with exceptionally fast inter-GPU communication (e.g., pure NVLink setups), applying only TP might be sufficient. However, SP remains essential for scaling large models across diverse hardware configurations where communication latency is a bottleneck.

\section*{Acknowledgments}
This work was supported by the New Faculty Startup Fund from Seoul National University, by the research fund of Hanyang University (HY-201700000002388), and by Automation and System Research Institute at Seoul National University (No. 0418-20250030). 
This work was also supported by Institute of Information \& communications Technology Planning \& Evaluation (IITP) grant funded by the Korea government (MSIT) (No. RS-2025-02214497, RS-2025-02263167, RS-2024-00438729, RS-2021-II211343, IITP-2026-RS-2021-II211817), and by the Basic Science Research Program through the National Research Foundation of Korea (NRF) funded by the Ministry of Education (RS-2026-25476387). 
We are grateful to VESSL AI for providing compute resources used in some of the early experiments in this work.
Jiwon Seo is the corresponding author.


\bibliography{custom}

\appendix

\section{Appendix}
\label{sec:appendix}
We will now show the derivations of the steady-state batch-size limits and the throughput gain of remaining-length-aware scheduling.

\textbf{A.1. Saturated Limit of Equation (1)}

Let the initial conditions be $s^a_0 = 0$, $s^r_0 = 0$, and $n_0 = B$, with $s^a_t = s^r_t = n_t = 0$ for $t < 0$. 
Let $B_{\mathrm{eff}}^{\star} = \lim_{t\to\infty} (s^a_t + n_t) = s^a + n$ be the steady-state effective bound. By the conservation of the total quantity in the system, $s^a_t + s^r_t + n_t = B$ for all $t$. 
Now:
$$B = B_{\mathrm{eff}}^{\star} + s^r$$
Where, by taking the limit $t \to \infty$ on the recurrence relation for $s^r_t$:
$$s^r = \sum_{i=1}^{N-1} (s^a + n)(1-\Theta) = (N-1)B_{\mathrm{eff}}^{\star}(1-\Theta)$$
Overall:
$$B_{\mathrm{eff}}^{\star} + (N-1)B_{\mathrm{eff}}^{\star}(1-\Theta) = B$$
$$B_{\mathrm{eff}}^{\star} = \frac{B}{N - \Theta(N-1)}$$
As desired.

\vspace{1em}

\textbf{A.2. Saturated Limit of Equation (3)}

Let the initial conditions be $B_i = B(1-\alpha)^i$ for $0 \le i \le N-1$. 
Let $B_{\alpha}^{\star} = \lim_{t\to\infty} B_t$. Rearranging the recurrence relation yields $B_{t+1} - B_t = \alpha (B_{t+1-N} - B_t)$. Summing both sides from $t = N-1$ to $T-1$ creates a telescoping sum.
Now:
$$B_T - B_{N-1} = \alpha \sum_{t=N-1}^{T-1} B_{t+1-N} - \alpha \sum_{t=N-1}^{T-1} B_t$$
Where, by shifting the index and canceling overlapping terms as $T \to \infty$:
$$B_{\alpha}^{\star} - B_{N-1} = \alpha \sum_{j=0}^{N-2} B_j - \alpha (N-1)B_{\alpha}^{\star}$$
And, using the initial condition for the geometric sum:
$$\alpha \sum_{j=0}^{N-2} B(1-\alpha)^j = B - B(1-\alpha)^{N-1} = B - B_{N-1}$$
Overall:
$$B_{\alpha}^{\star} - B_{N-1} = (B - B_{N-1}) - \alpha(N-1)B_{\alpha}^{\star}$$
$$B_{\alpha}^{\star} + \alpha(N-1)B_{\alpha}^{\star} = B$$
$$B_{\alpha}^{\star} = \frac{B}{1 + \alpha(N-1)}$$
As desired.

\textbf{A.3. Derivation of Throughput Ratio G}

We derive \(G\) over the period between two consecutive request
arrivals. Following the notation in
Section~\ref{sec:length-scheduling}, we further let
\(\mathrm{Tok}_x\) and \(T_x\) denote the decoded-token count and
execution time for \(x\in\{\textit{aggr},\textit{delay}\}\),
respectively.

\begingroup
\setlength{\jot}{2pt}
\[
\begin{aligned}
\mathrm{Tok}_{\mathrm{aggr}}
&= (bs+1)I-\sum_{k=1}^{n} i_{\mathrm{wait}}^{k}, \\
T_{\mathrm{aggr}}
&= \bigl(I-(n+1)\bigr)t_d+(n+1)t_p, \\
\mathrm{Thr}_{\mathrm{aggr}}
&= \frac{\mathrm{Tok}_{\mathrm{aggr}}}
         {T_{\mathrm{aggr}}}, \\[2pt]
\mathrm{Tok}_{\mathrm{delay}}
&= (bs+1)I-i_{\mathrm{delay}}, \\
T_{\mathrm{delay}}
&= (I-1)t_d+t_p, \\
\mathrm{Thr}_{\mathrm{delay}}
&= \frac{\mathrm{Tok}_{\mathrm{delay}}}
         {T_{\mathrm{delay}}}, \\[2pt]
G
&= \frac{\mathrm{Thr}_{\mathrm{delay}}}
         {\mathrm{Thr}_{\mathrm{aggr}}}.
\end{aligned}
\]
\endgroup

\end{document}